\documentclass[review]{elsarticle}
\graphicspath{ {./figures/} }
\usepackage{hyperref}
\usepackage{float}
\usepackage{verbatim} 
\usepackage{apalike}
\restylefloat{figure}
\restylefloat{table}
\usepackage{algorithm}
\usepackage{algpseudocode}
\usepackage{xcolor}
\usepackage{tabularx} 
\usepackage{amsmath}
\usepackage{adjustbox}
\usepackage{booktabs} 

\usepackage[numbers]{natbib}
\usepackage{listings}
\definecolor{codegreen}{rgb}{0,0.6,0}
\definecolor{codegray}{rgb}{0.5,0.5,0.5}
\definecolor{codepurple}{rgb}{0.58,0,0.82}
\definecolor{backcolour}{rgb}{0.95,0.95,0.92}
\lstdefinestyle{mystyle}{
    backgroundcolor=\color{backcolour},   
    commentstyle=\color{codegreen},
    keywordstyle=\color{magenta},
    numberstyle=\tiny\color{codegray},
    stringstyle=\color{codepurple},
    basicstyle=\ttfamily\footnotesize,
    breakatwhitespace=false,         
    breaklines=true,                 
    captionpos=b,                    
    keepspaces=true,                 
    numbers=left,                    
    numbersep=5pt,                  
    showspaces=false,                
    showstringspaces=false,
    showtabs=false,                  
    tabsize=2
}

\journal{arXiv}

\begin{document}
\begin{frontmatter}

\title{COGNET-MD, an evaluation framework and dataset for Large Language Model benchmarks in the medical domain}


\author[label1]{Dimitrios P. Panagoulias, MSc.CS}
\ead{panagoulias\_d@unipi.gr}
\author[label2]{Persephone Papatheodosiou, MD-MSc}
\author[label3]{Anastasios P. Palamidas, MD-PhD}
\author[label4]{Mattheos Sanoudos, DDS-PhD}
\author[label2]{Evridiki Tsoureli-Nikita, MD-PhD}
\author[label1]{Maria Virvou, PhD}

\author[label1, cor1]{George A. Tsihrintzis, PhD}

\cortext[cor1]{Corresponding author.}
\address[label1]{Department of Informatics, University of Piraeus, Piraeus 185 34, Greece}
\address[label2]{School of Medicine, National and Kapodistrian University of Athens, Athens, Greece}
\address[label3]{Interventional Pulmonology Department, Athens Medical Center, Greece}
\address[label4]{School of Dentistry, Department Of Orthodontics, National and Kapodistrian University of Athens, Athens, Greece}

\begin{abstract}

Large Language Models (LLMs) constitute a breakthrough state-of-the-art Artificial Intelligence (AI) technology which is rapidly evolving and promises to aid in medical diagnosis either by assisting doctors or by simulating a doctor's workflow in more advanced and complex implementations. In this technical paper, we outline Cognitive Network Evaluation Toolkit for Medical Domains (COGNET-MD), which constitutes a novel benchmark for LLM evaluation in the medical domain. Specifically, we propose a scoring-framework with increased difficulty to assess the ability of LLMs in interpreting medical text. The proposed framework is accompanied with a database of Multiple Choice Quizzes (MCQs). To ensure alignment with current medical trends and enhance safety, usefulness, and applicability, these MCQs have been constructed in collaboration with several associated medical experts in various medical domains and are characterized by varying degrees of difficulty. The current (first) version of the database includes the medical domains of Psychiatry, Dentistry, Pulmonology, Dermatology and Endocrinology, but it will be continuously extended and expanded to include additional medical domains.

\end{abstract}

\end{frontmatter}

\section{Problem Statement}

Large Language Models (LLMs) are advanced computational algorithms designed to generate and manipulate natural language, when triggered by a human prompt. At the core of these models lie a subset of machine learning models that have been trained on vast amounts of text, audio, image, and video data and possess the ability to generate knowledge, reason on specific prompts and provide a conversational agent that can enhance human abilities on a multitude of tasks \cite{gpt44, devlin2018bert, agarwal2000time}. To increase LLM efficiency and AI safety, many techniques and methods have been developed and investigated \cite{khan2022transformers}. For each method, suitability is dependent on task, budget and time. However, in any deployment or use case and whether or not precision and usability are the main parameters influencing development decisions, evaluation metrics on accuracy and specificity on the generative abilities of LLMs are of highest importance \cite{panagoulias2024augmenting, panagoulias2024dermacen, panagoulias2024evaluating}. 

The medical domain is a field that can greatly benefit from use of AI-empowered applications, including LLMs. Indeed, if properly developed, evaluated and used, such application can provide significant assistance to medical and healthcare professionals \cite{blandford2020opportunities,snoswell2023clinical}. Unfortunately, currently, there appears to be a shortage of independent datasets available for the evaluation of LLMs in the medical field. This, in turn, creates barriers and friction in the deployment of medical AI-empowered applications.

\section{COGNET-MD Version 1.0}

To address the lack of or the limited availability of independent, free to use and ready to assess datasets for LLM evaluation in the medical domain \cite{gilson2023does},  we have constructed Cognitive Network Evaluation Toolkit for Medical Domains (COGNET-MD). Moreover, to improve deployment and usage, we also propose use cases of varying difficulty for its accompanying dataset. More specifically, the dataseth consists of 542 datapoints of domain-specific questions with one or more correct choices/answers. In table~\ref{tab:Cog} (in the Appendix), a preview of COGNET-MD can be seen. The complete dataset can be found and loaded for evaluation on HuggingFace \footnote{https://huggingface.co/datasets/DimitriosPanagoulias/COGNET-MD/} and in Version 1 includes MCQs about Dentistry \cite{hawley2009dental, bagheri2013clinical,mccann2010teaching}, Dermatology \cite{bolognia2014dermatology, james2019andrews, goldsmith2012fitzpatrick}, Endocrinology \cite{williams2016,gardner2015, grossman2010,fadem2010,fenichel2016}, Psychiatry \cite{Sadock2015,Gabbard2014,First2014, Stahl2013,Tasman2015} and Pulmonology \cite{ditki_respiratory,statpearls_pediatric,ers_hermes,abp_pediatric}.  

\begin{figure}
    \centering
    \includegraphics[width=1\linewidth]{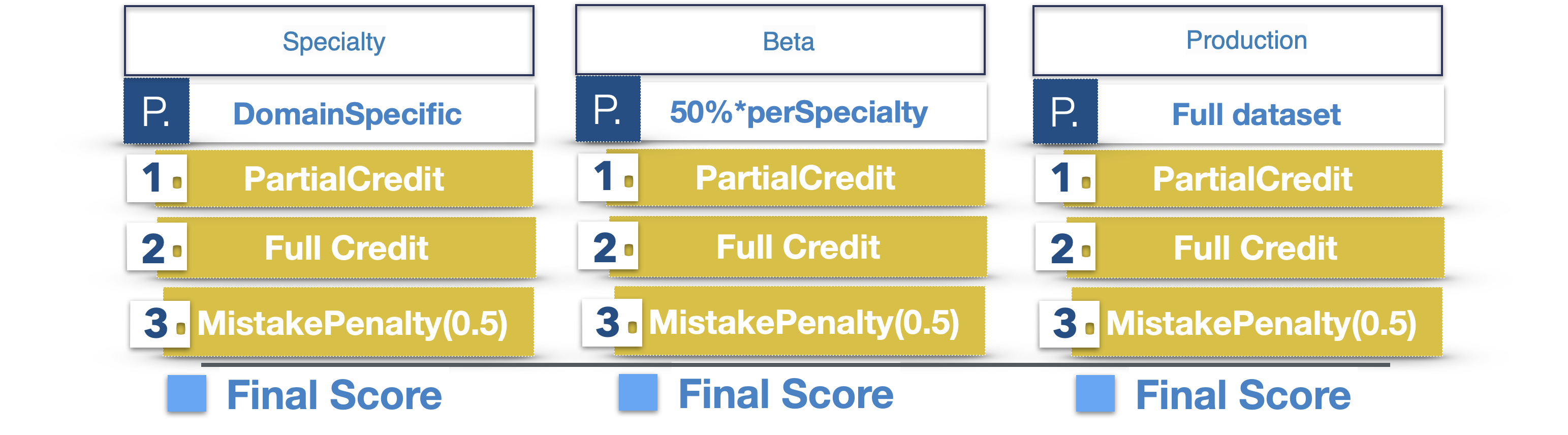}
    \caption{Benchmark Varying difficulty use Cases}
    \label{fig:bench}
\end{figure}
In figure~\ref{fig:bench}, the benchmark use cases are presented as Specialty where one medical Specialisation is chosen, Beta where 50\% per specialty of total dataset is chosen with all included medical specialties and Production where all Dataset is used. 
The scoring for each use case is the same and includes the following as shown in figure \ref{fig:bench} and the Algorithm 1:
\begin{itemize}
\item Partial Credit: At least one correct answer equals to a half point - 0.5. 
\item Full Credit: To achieve full points depending on difficulty either all correct answers must be selected and no incorrect ones or a correct response gets the full credit, equals to 1 point.
\item Penalty for Incorrect Answers: Points are deducted for any incorrect answers selected. -(minus) 0.5 point for each incorrect answer selected.
\end{itemize}

The dataset can be used to assess the model's ability to infer relationships between specialties and knowledge spaces. Thus it can be analyzed either as a whole, encompassing all included specialties-full Dataset, partially or it can be narrowed down to focus on a specific medical domain-specialty. A rule based algorithm to be used as a deployment paradigm is presented below: 

\underline{\textbf{Algorithm 1 Scoring Algorithm Based on Difficulty Level}}
\label{algo}
\begin{algorithmic}[1]
\Function{ScoreResponses}{user\_responses, correct\_responses}
        \ForAll{(user\_response, correct\_response) \textbf{in} zip(user\_responses, correct\_responses)}
            \State correct\_selected $\gets$ user\_choices $\cap$ correct\_choices
            \State incorrect\_selected $\gets$ user\_choices - correct\_choices
            
            \If{correct\_selected}
                \State score $\gets$ score + 0.5
            \EndIf
            
            \If{correct\_selected == correct\_choices \textbf{and} not incorrect\_selected}
                \State score $\gets$ score + 0.5
            \EndIf
            
            \State score $\gets$ score - 0.5 * \textbf{len}(incorrect\_selected)
        \EndFor
        \State \Return score
\EndFunction
\end{algorithmic}
In the following section the Code of Conduct as a space for a score to be considered valid and the rules of conduct as one-shot and few shot prompt examples, to be used for each use case are presented.

\section{Rules, Code of Conduct and Prompt Examples}
For a score to be valid and be added in the COGNET-MD's leader-boards the developers, should clearly state model used, add a short model description and use case scenario used, as described in the previous section. In the following Benchmark Card two examples are presented:

\begin{table}[H]
    \centering
    \begin{tabular}{ccccc}
       MODEL  & Description & Domain &Difficulty  & COGNET-MD VERSION\\
       \hline
       GPT-4  & Add here related info & Full Dataset & Production & 1.0 \\
       Mistral & Add here related info & Dermatology & Specialty & 1.0 \\
    \end{tabular}
    \caption{Benchmark Card}
    \label{tab:my_label}
\end{table}

If for any reason the developer chooses not to use the complete dataset as per content, then the developer should clearly state it.

\textbf{One Shot example}

\begin{lstlisting}[language=Python]
rule_of_conduct = ``On a Multiple choice Quiz choose correct responses:(ONLY THE CORRECT LETTERS and no spaces and no other associated Text. If more than one letter then add a dash- between letters)."
MODEL = ``MODEL''
\end{lstlisting}
In this scenario if one domain is chosen it can be clearly stated in the prompt as it has shown to slightly increase accuracy \cite{panagKes}. So if Dermatology is chosen then the prompt would be:

\textbf{Few Shot example Domain Specific}
\begin{lstlisting}[language=Python]
rule_of_conduct = "On the Following Dermatology Multiple choice Quiz choose correct response or responses. This is an example question: Which condition may include generalised pruritus as a symptom? A. Hodgkins disease B. Pityriasis rosea C. Diabetes mellitus D. Haemolytic jaundice E. Polycythaemia rubra vera and the Correct responses on that example are A-B-E."
MODEL = ``MODEL''
\end{lstlisting}

\section*{Acknowledgements}
This work has been partly supported by the University of Piraeus Research Center.

\section{Appendix}
\begin{table}[H]
    \centering
    \caption{COGNET-MD evaluation dataset}
    \begin{tabular}{p{9.0cm}cc}        
    \toprule
         Questions & Correct-Choices & Specialty \\
         \midrule \hline
         In which neurological condition is Gowers' sign or maneuver typically observed?  A. Multiple sclerosis B. Myotonic dystrophy C. Huntington’s disease D. Duchenne’s muscular dystrophy E. Myasthenia gravis & D & Psychiatry \\ \hline
         A husband and wife present at an emergency room. The wife informs that her husband has developed an unwavering belief of infidelity on her part, which she denies. His behaviors include tailing her, sniffing her clothes, scrutinizing her purse, and consistently accusing her. He does not fit the mood disorder criteria and denies having other psychotic symptoms. His medical history displays no substance abuse. What could be his diagnosis?  A. Schizophrenia B. Major depressive disorder with psychotic features. C. Delirium D. Dementia E. Delusional disorder& E & Psychiatry\\ \hline
        Which among the following options might prove beneficial for an old man suffering from a gravitational ulcer with adjacent eczema?  A. The usage of a supportive elastic bandage B. Daily application of Betnovate-C ointment C. Povidine-iodine ointment on the ulcer D. Diuretic therapy for oedema reduction E. Lassar's paste application around the ulcer & A-C-E & Dermatology\\ \hline
        Which condition may include generalised pruritus as a symptom?  A. Hodgkin's disease B. Pityriasis rosea C. Diabetes mellitus D. Haemolytic jaundice E. Polycythaemia rubra vera & A-B-E & Dermatology \\ \hline 
    \end{tabular}
    \caption{COGNET-MD evaluation dataset}
    \label{tab:Cog}
\end{table}
\section*{Data Availability}
Data and python code can be found at:
\url{https://huggingface.co/datasets/DimitriosPanagoulias/COGNET-MD/}, \url{https://huggingface.co/datasets/DimitriosPanagoulias/COGNET-MD/blob/main/COGNET-MD1.0.ipynb}. Use the Benchmark Card to be added on or future leader boards of COGNET-MD. This is version 1.0 and more data will be added shortly. If you want to contribute to the database send your data for consideration and expert evaluation at the corresponding authors 
\label{availability}


\section*{Abbreviations}
The following abbreviations are used in this paper:\\
\begin{tabular}{@{}ll}
LLM & Large Language Model \\
AI & Artificial Intelligence \\
MCQ & Multiple-choice Questionnaire \\
\end{tabular}

\bibliography{ref2}

\end{document}